\pdfoutput=1

\documentclass[11pt]{article}

\usepackage{arxiv_silveralign}

\usepackage{times}
\usepackage{latexsym}
\usepackage{tabularx}
\usepackage{makecell}
\usepackage{multirow}
\usepackage{xspace}
\usepackage{scrextend}
\usepackage[inline]{enumitem}
\usepackage{tikz}
\usepackage{float}
\usepackage{booktabs}
\usepackage{algorithm}
\usepackage{algpseudocode}
\usepackage{amsmath}
\usepackage{threeparttable}

\usepackage[T1]{fontenc}

\usepackage[utf8]{inputenc}

\usepackage{microtype}

\usepackage{inconsolata}

\newcounter{notecounter}
\newcommand{\enotesoff}{\long\gdef\enote##1##2{}}
\newcommand{\enoteson}{\long\gdef\enote##1##2{{
			\stepcounter{notecounter}
			{\large\bf
				\hspace{1cm}\arabic{notecounter} $<<<$ ##1: ##2
				$>>>$\hspace{1cm}}}}}

\newcommand{\precision}{\operatorname{precision}}
\newcommand{\recall}{\operatorname{recall}}
\newcommand{\fone}{\operatorname{F_1}}

\enoteson
\enotesoff

\def\ourmethod{SilverAlign\xspace}

\title{\ourmethod: MT-Based Silver Data Algorithm For Evaluating Word Alignment}

\author{First Author \\
	Affiliation / Address line 1 \\
	Affiliation / Address line 2 \\
	Affiliation / Address line 3 \\
	\texttt{email@domain} \\\And
	Second Author \\
	Affiliation / Address line 1 \\
	Affiliation / Address line 2 \\
	Affiliation / Address line 3 \\
	\texttt{email@domain} \\}

\author{Abdullatif Köksal\textsuperscript{1,2}, Silvia Severini\textsuperscript{1}, Hinrich Schütze\textsuperscript{1}\\
	\textsuperscript{1}Center for Information and Language Processing (CIS), LMU Munich, Germany \\
	\textsuperscript{2}Munich Center for Machine Learning (MCML), Germany \\
	\texttt{\{akoksal, silvia\}@cis.lmu.de} }

\begin{document}
	\maketitle
	\begin{abstract}
		
		Word alignments are essential for a variety of NLP tasks. Therefore, choosing the best approaches for their creation is crucial. However, the scarce availability of gold evaluation data makes the choice difficult. 
		We propose \ourmethod, a new method to automatically create silver data for the evaluation of word aligners by exploiting machine translation and minimal pairs. 
		We show that performance on our silver data
		correlates well with gold benchmarks for 9 language
		pairs, making our approach a valid resource for
		evaluation of different domains and languages when
		gold data are not available. This
		addresses the important scenario of missing gold data alignments for
		low-resource languages.
	\end{abstract}

	\section{Introduction}

	Word alignments (WA) are crucial for statistical machine translation (SMT) where they constitute the basis for creating probabilistic translation dictionaries. They are relevant to different tasks such as neural machine translation (NMT) \citep{alkhouli-etal-2018-alignment}, typological analysis \citep{ostling-2015-word}, annotation projection \citep{huck-etal-2019-cross}, bilingual lexicon extraction \cite{mckeown1996translating,ribeiro2001extracting}, and for creating multilingual embeddings \citep{dufter-etal-2018-embedding}. Different approaches have been investigated using statistics like IBM models \citep{brown1993stephen} and Giza++ \citep{och2003systematic}. 
	More recently, \citet{ostling2016efficient}  introduced Eflomal, a high-quality word aligner widely used nowadays for its ability to align many languages effectively.
	Other methods create alignments from attention matrices of NMT models \citep{zenkel2019adding}, solve multitask problems \citep{garg-etal-2019-jointly}, or leverage multilingual word embeddings \citep{sabet2020simalign}.

	\begin{figure}[t]
		\centering
		\includegraphics[width=0.48\textwidth]{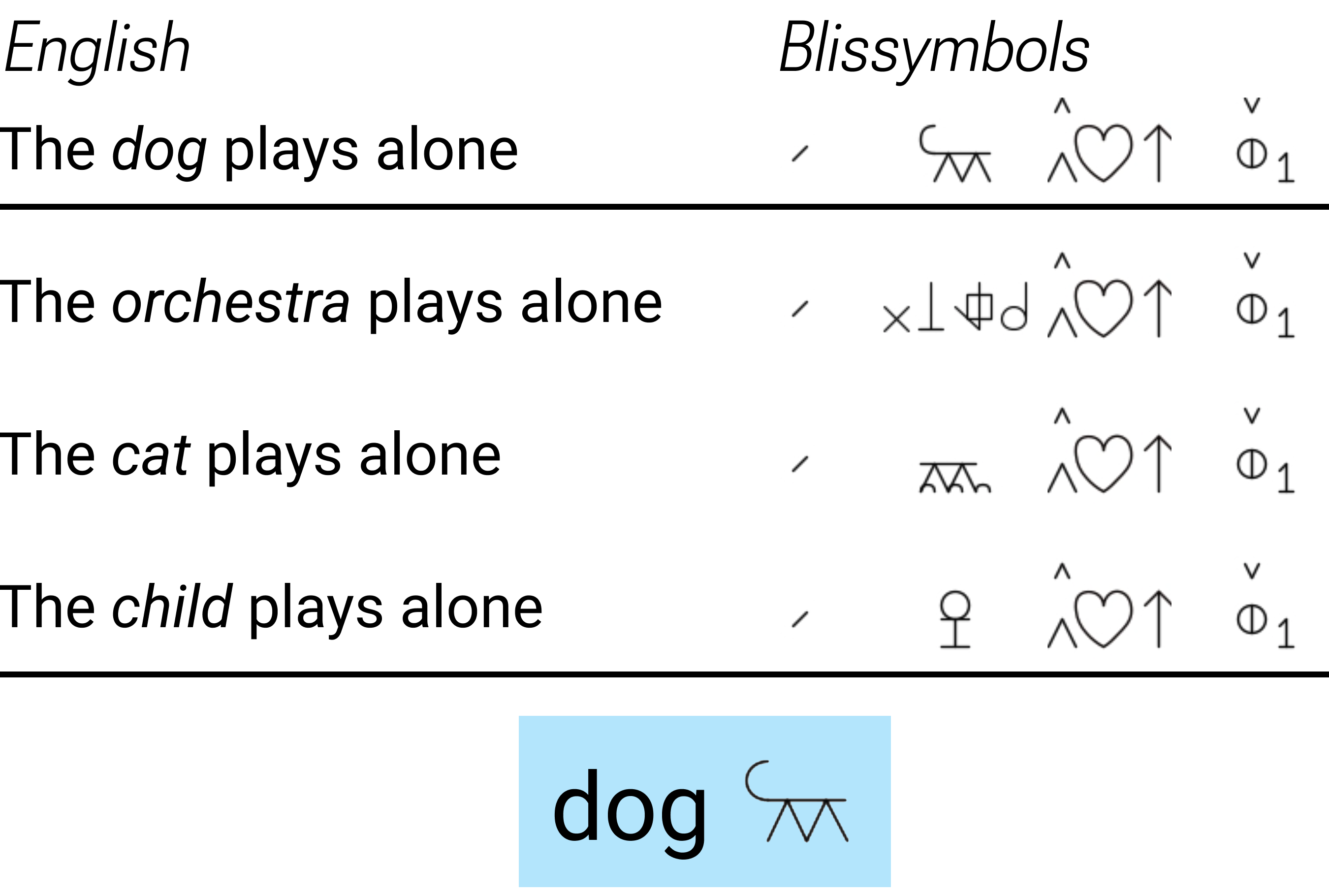}
		\caption{An example of our technique with minimal pairs for a source sentence in English and the target language, Blissymbols.\footnotemark  For a word (dog) in the source sentence, we create minimal pairs (orchestra, cat, child), and then we can align the word dog to the correct symbol in Blissymbols with the help of translations.}
		\label{fig:example}
	\end{figure}

	Given the variety of approaches available for aligning words, the choice of the best alignment methods for a certain parallel corpus has gained attention.
	Such decision requires evaluation data for the pair of languages and specific domain addressed. 
	Collecting gold data or high-quality word alignment benchmarks requires the work of various annotators as for the Blinker project of \citet{melamed1998manual} and WA shared tasks \citep{mihalcea2003evaluation,martin2005word} which can be a time-consuming or impractical job for lesser spoken languages. \citet{melamed1998manual} reports that annotating word alignments for 100 sentences in English-French would take 9 to 22 hours. Additionally, the annotation process often leads to conflicts among annotators \citep{macken2010annotation}. Hence, gold data is scarce or completely unavailable for many low-resource languages
	and, when dealing with domain-specific data such as medical or legal text, such availability is even less.
	Therefore analyzing existing word alignment models with a varying number of language pairs in different domains is a challenging task.

	\footnotetext{Blissymbols is a constructed language
		established in 1949 to help people with
		communication difficulties. The \hbox{\url{blissonline.se}} dictionary is used for the examples.}
	
	We propose \ourmethod, a novel algorithm to create silver evaluation data for guiding the choice of appropriate word alignment methods. 
	Our approach is based on a machine translation model
	and exploits minimal sentence pairs to create
	parallel corpora with alignment
	links. Figure \ref{fig:example} illustrates our core
	idea with minimal pairs in English and Blissymbols.
	Our approach is to create
	alternative sentences in minimal pairs, to rely on machine translation models to track changed words for each alternative and then align words in the source sentence.

	In summary, our contributions are:
	\begin{enumerate}
		\item We find that
		our silver benchmarks 
		rank methods with high consistency compared to rankings
		based on gold data. This means that we can identify the best
		methods based on silver data if there is no gold data
		available, which is frequently the case in low-resource
		scenarios for word alignment.
		
		\item We conduct an extensive analysis of our silver
		resource with respect to gold data for 9 language pairs from different language
		families and resource availability. We perform
		various experiments for word alignment models on
		sub-word tokenization, tokenizer vocabulary size,
		varying performance of Part-of-Speech tags, and word
		frequencies. 
		\item SilverAlign supports a more accurate evaluation
		and a more in-depth analysis than small gold sets (i.e., English-Hindi has only 90 sentences) because we can automatically create
		larger evaluation benchmarks. Also, \ourmethod is robust to domain changes as it shows a high correlation between gold and both in- and out-of-domain silver benchmarks.
		\item It has been shown that machine translation performance
		(including NMT performance) can be
		improved by choosing a tokenization  that optimizes
		compatibility between source and target languages  \cite{deguchi2020bilingual}. We show that \ourmethod
		can be used to find such a compatible tokenization for each language pair.
		\item We make our silver data and code available as a
		resource for future work that takes advantage of our silver evaluation datasets.\footnote{\url{https://github.com/akoksal/SilverAlign}} Our code can be used to create silver benchmarks for multiple languages, and our silver benchmark can be used out-of-the box.

	\end{enumerate}

	\enote{hs}{you did not do a good job at clearly describing
		and properly higlighting the contributions of the
		paper. here is a possible (probably partial) list.
		(*) most importantly, we find that
		our silver benchmarks 
		rank methods with high consistency compared to rankings
		based on gold data. this means that we can identify the best
		methods based on silver data if there is no gold data
		available, which is frequently the case in low-resource
		scenarios.
		(*) It has been shown that machine translation performance
		(including neural machine translatino perforeamnce) can be
		improved by choosing a tokenization  that optimizes
		compatibility between source and target languages. Our method
		can be used ot find such a compatible tokenziation. (cite
		https://aclanthology.org/2020.coling-main.378/ ?)
		(*) extensive experimentation with automatically created
		silver datasets on X different tasks, demonstrating the
		effectiveness of SilverAlign.
		(*) SilverAlign supports more accurate evaluation results
		and more in-depth analysis because we can automatically create
		large evaluation benchmarks.
		(*) robust against domain change?
		(*) we publish code and data
		(*) Silver data are made available: this is a useful
		resource for evaluation EVEN IF SOMEBODY DOES NOT WANT TO
		USE
		THE SilverAlign DATA CREATION METHODOLOGY: THEY CAN SIMPLY
		DIRECTLY USE THE SILVER STANDARDS WE CREATED
	}
	
	\enote{hs}{if you have space, it would be nice to include an
		overview of the sections of the paper here}

	The rest of the paper is organized as follows. Section \ref{related_work} describes 
	related work. The
	details of \ourmethod method are explained in Section
	\ref{method}. Section \ref{experimental_setup} describes the experimental setup, evaluation metrics and 
	datasets. We compare the results on
	our silver benchmarks to gold data in Section \ref{results}.
	Finally, we draw conclusions and
	discuss future work in Section \ref{conclusion}.

	\section{Related Work}
	\label{related_work}
	
	\subsection{Word alignment analysis}
	The analysis of word alignment performance with respect to different factors has been analyzed by many works.
	\citet{ho2019neural} compare discrete and neural version word aligners and show the superiority of the second class.	They also compare them with respect to unaligned words, rare words, Part-of-Speech (PoS) tags, and distortion errors.
	\citet{asgari2020subword} study word alignment results when using subword-level tokenization and show improved performance with respect to word level.
	\citet{sabet2020simalign} analyze the performance of
	word aligners regarding different PoS for
	English/German and show that Eflomal has low
	performance when aligning links with high distortion. They also analyze the alignments based on word frequency and show that the performance decreases for rare words when aligning at the word level versus the subword level.
	
	\citet{ho2021optimizing} analyze the interaction between alignment methods and subword tokenization (Unigram and Byte Pair Encoding (BPE)). They observe that tokenizing into smaller units helps to align rare and unknown words. They also investigate the effect of different vocabulary sizes and conclude that word-based segmentation is less optimal. We also conduct an experiment in this direction in Section \ref{sec:tokenizer_vocabulary_size}.
	
	\begin{figure*}[!ht]
		\centering
		\includegraphics[width=\textwidth]{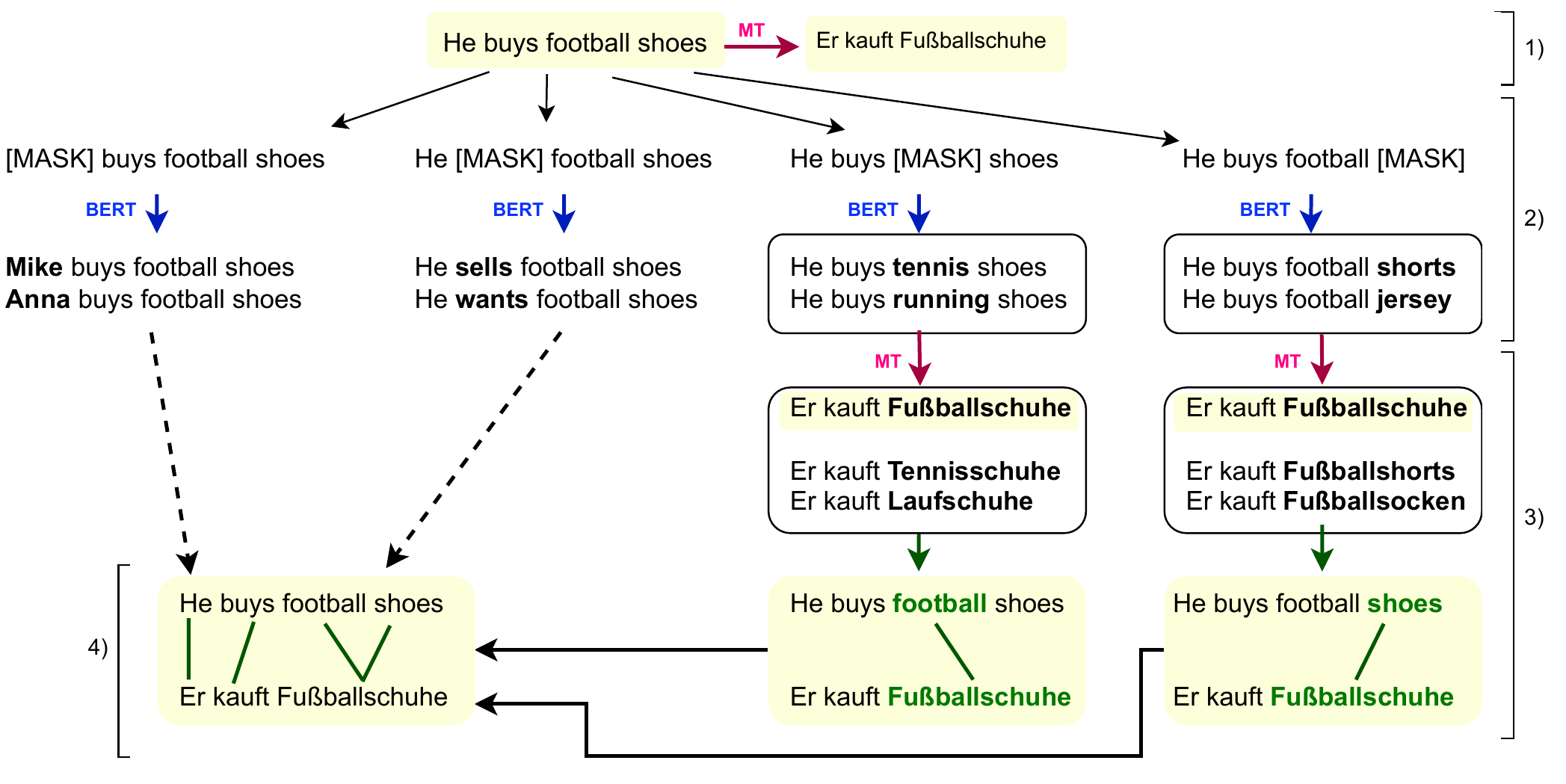}
		\caption{One-to-one and many-to-one examples of English-German alignments according to \ourmethod.
			The first step translates a
			given source sentence ($s_i$) to a
			sentence ($t_i$) in the target
			language via machine
			translation. The second step finds
			alternatives for each word in $s_i$
			via a foundation model trained
			with the masked language modeling
			(MLM) objective. The third step
			translates all alternative sentences
			of a given word and tracks which
			words are changed in $t_i$. If
			alternative translations change the
			same word in $t_i$, we label the
			corresponding alignments and merge
			the affected words in the fourth step. (Details for the first two columns omitted for clarity). 
		}
		\label{fig:pipeline}
	\end{figure*}
	
	\subsection{Silver data creation in NLP} 
	Collecting gold data for evaluating or training systems can be impractical due to its cost and the need for human annotators. To solve these issues, silver data - data generated automatically - has been widely exploited for different tasks and domains.
	For the Named Entity Recognition (NER) task, \citet{rebholz2010calbc}  introduce CALBC, a silver standard corpus  generated by the harmonization of multiple annotations, \citet{wu2021unitrans} create training data for their NER model through word-to-word machine translation and annotation projection, and \citet{severini2022towards} create named entities pairs from co-occurrence statistics and transliteration models.
	For the medical domain, there exist multiple silver sets due to the difficulty of finding qualified annotators. Examples are the silver corpus of \citet{rashed2020english} for training and evaluating COVID-19-related NLP tools, and DisTEMIST from \citet{miranda2022overview}, a multilingual dataset for 6 languages created through annotation transfer and MT for automatic detection and normalization of disease mentions from clinical case documents. 
	\citet{paulheim2013dbpedianyd} introduced DBpedia-NYD for evaluating the semantic relatedness of resources in DBpedia and exploiting web search engines.
	\citet{baig2021towards} propose a silver-standard dependency treebank of Urdu tweets using self-training and co-training to automatically parse big amounts of data. 
	\citet{wang2022expanding} synthesize labeled data using lexicons to adapt pretrained multilingual models to low-resource languages.

	\section{Method}
	\label{method}

	\begin{table*}[!ht]
		\small
		\centering
		\begin{threeparttable}
			\begin{tabular}{l|rr|rr}
				\toprule
				
				Lang & Gold data & Size & Parallel data & Size \\
				\midrule
				ENG-CES &     \citet{marevcek2008automatic}      &     2501           &       EuroParl \citep{koehn2005europarl}       &      648K          \\   
				ENG-DEU &  EuroParl-based\tnote{a}         &      508    &     EuroParl \citep{koehn2005europarl}     &       1907K         \\   
				ENG-FAS &      \citet{tavakoli2014phrase}     &      400   &     TEP \citep{pilevar2011tep}          &         595K       \\   
				ENG-FRA &     WPT2003, \citet{och2000improved}      &       447    &     Hansards \citep{germann2001aligned}          &     1123K           \\   
				ENG-HIN &    WPT2005\tnote{b}  &         90  &      Emille \citep{mcenery2000emille}        &     3K           \\   
				ENG-RON &     WPT2005\tnote{b}
				
				&       199    &        Constitution, Newspaper \tnote{b}
				
				&        39K        \\   
				ENG-TUR &    PBC-based (Our)      & 100 &     PBC \citep{mayer-cysouw-2014-creating}          &    30K            \\   
				\midrule
				FIN-ELL &      \citet{yli-jyra-etal-2020-helfi}     &   7,909             &      PBC \citep{mayer-cysouw-2014-creating}     &   8K             \\   
				FIN-HEB &     \citet{yli-jyra-etal-2020-helfi}     &       22,291         &     PBC \citep{mayer-cysouw-2014-creating}    &   22K               \\   
				\bottomrule
			\end{tabular}
			\begin{tablenotes}
				\item[a]  \url{www-i6.informatik.rwth-aachen.de/goldAlignment/} 
				\item[b] \url{http://web.eecs.umich.edu/~mihalcea/wpt05/}
			\end{tablenotes}
		\end{threeparttable}
		
		\caption{Overview of gold and parallel datasets. ``Size'' refers to the number of sentences. Language pairs are represented with their respective ISO 639-3 codes.}
		\label{tab:data}
	\end{table*}

	\begin{table}[!ht]
		\centering
		\resizebox{.48\textwidth}{!}{
			\begin{tabular}{l|rr|rr|rr}
				\toprule
				\multirow{2}{*}{Lang} & \multicolumn{2}{c|}{Gold} & \multicolumn{2}{c|}{Silver\textsubscript{Small}} & \multicolumn{2}{c}{Silver\textsubscript{Large}}\\ 
				& Size & $|A|$ & Size & $|A|$  & Size & $|A|$ \\
				\midrule
				ENG-CES &    2,501 & 67K   &    1,507 & 3,852   &    26K & 57K        \\   
				ENG-DEU &    508 & 11K   &    227 & 480  &    31K & 77K  \\   
				ENG-FAS &    400 & 12K  &    137 & 242   &    16K & 27K         \\   
				ENG-FRA &    447 & 17K   &    216 & 359   &    32K & 74K            \\   
				ENG-HIN &    90 & 1,409   &    46 & 87  &    26K & 58K   \\   
				ENG-RON &    199 & 5,034   &    69 & 161   &    28K & 64K  \\
				ENG-TUR &    100 & 2,670   &    50 & 80   &    27K & 60K             \\   
				\midrule
				FIN-ELL  &    7909 & 161K & 1,668   &    2,230 & - &  -    \\   
				FIN-HEB &   22,291  & 405K   &    4,522 & 6,396  &    - & -           \\   
				\bottomrule
			\end{tabular}
		}
		\caption{Overview of gold and silver dataset sizes. Size refers to the number of sentences and $|A|$ is the total number of alignments.}
		\label{tab:silver}
	\end{table}

	The pipeline of our silver data creation algorithm is illustrated in Figure \ref{fig:pipeline}. 
	Given a source language $S$ and a target language $T$, we now describe the steps to create our word alignment silver data for $S$-$T$:
	
	\begin{enumerate}
		\item 
		We first collect monolingual data from the source language, $D_{S}$. 
		Given a sentence $s_i={w^{s}_1, w^{s}_2, ..., w^{s}_N} \in D_S$ of length $N$, we use a machine translation system to generate the target sentence $t_i={w^{t}_1, w^{t}_2, ..., w^{t}_M}$, and therefore target data $D_{T}$. 
		
		\item Then, we create minimal pairs for $s_i$ by finding alternative words for each $w^{s}_j$ in the sentence ($j\in[1,N])$. 
		We use a pretrained Masked Language Model (i.e., English BERT\textsubscript{Large}) to find alternative words which fit into the context well. For each $s_i$, we create five alternatives per word by masking one word at a time. Examples of minimal pairs for the sentence ``I love pizza'' are ``You love pizza'', ``I hate pizza'', and ``I love apples''.
		
		\item In the third step, we use a machine
		translation system to translate all alternative
		sentences to the target language.\footnote{As recent
			machine translation models \cite{nllb} support the
			pairwise translation of more than 200 languages,
			our method is applicable to several hundred languages and
			thousands of language pairs.}
		Based on the changed words in the alternative sentences,	
		we 
		
		align the words in 
		$s_i$ to 
		the words in $t_i$. For example, given a sentence of
		length $4$ as in Figure \ref{fig:pipeline} and a
		masked word in position $1$, if the alternatives
		present the same $w^{t}_2$ (\textit{kauft}) and
		$w^{t}_3$ (\textit{Fußballschuhe}), but words different from $w^{t}_1$ (\textit{Mike, Anna}), then $w^{t}_1$ (\textit{Er}) is aligned to the masked word, $w^{s}_1$ (\textit{He}).
		
		The method works under the following three assumptions.
		First, each alternative of a word is valid as long as the length of the translation to the target language stays the same. 
		Second, at least 4 out of 5 alternative words should be valid.
		Third, all the valid alternatives should change the
		same word and only one word in the target language, consistently. 
		Even though these constraints
		mean that our ``yield'' is low (many words are
		not aligned because they do not satisfy the constraints),

		they are necessary to get confident word alignments. 
		As collecting monolingual data is easy, we overcome this size problem by using a larger monolingual corpus (see \ref{sec:silver_data_size}).

		\item Finally, we merge the alignment links for each
		$w_j\in s_i$ to create
		a set of silver alignments for the sentence pair $s_i$-$t_i$.
	\end{enumerate}
	
	\section{Experimental Setup}
	\label{experimental_setup}

	Table \ref{tab:data} shows the gold and parallel data sources and statistics. The chosen set of languages represents diverse language families, scripts, and resource availability. 
	In order to include a challenging benchmark containing a dissimilar language pair, we propose a new gold dataset for English-Turkish (a language with poor morphology and a highly agglutinative language with complex morphology). We first collect 100 random English sentences from the Parallel Bible Corpus of \citet{mayer-cysouw-2014-creating} and translate them to Turkish with Google Translate. Then, gold word alignments are annotated by a native speaker annotator who is also in charge of fixing any translation issues.
	
	\citet{yli-jyra-etal-2020-helfi} present a gold dataset for Finnish and Ancient Greek. However, 
	we were not able to find a machine translation model from Finnish to Ancient Greek so we compare the Finnish - Modern Greek silver data to the Finnish - Ancient Greek gold data.
	
	In all our experiments, we use Google Translate as our machine translation model. For Hebrew, the model produces words without vowels since this language is standardly written without them. However, this is not the case for \citep{yli-jyra-etal-2020-helfi}'s gold dataset,  so we pre-process the latter by removing short vowels.
	We use Eflomal \citep{ostling2016efficient} with the grow-diag-and-final (GDFA) symmetrization method as a word aligner to compare different configurations such as tokenizers. 
	
	To create alternative sentences, we use English BERT\textsubscript{Large} and Finnish BERT \citep{virtanen-etal-2019-finbert}.
	
	\begin{figure*}[ht!]
		\centering
		\includegraphics[width=\textwidth]{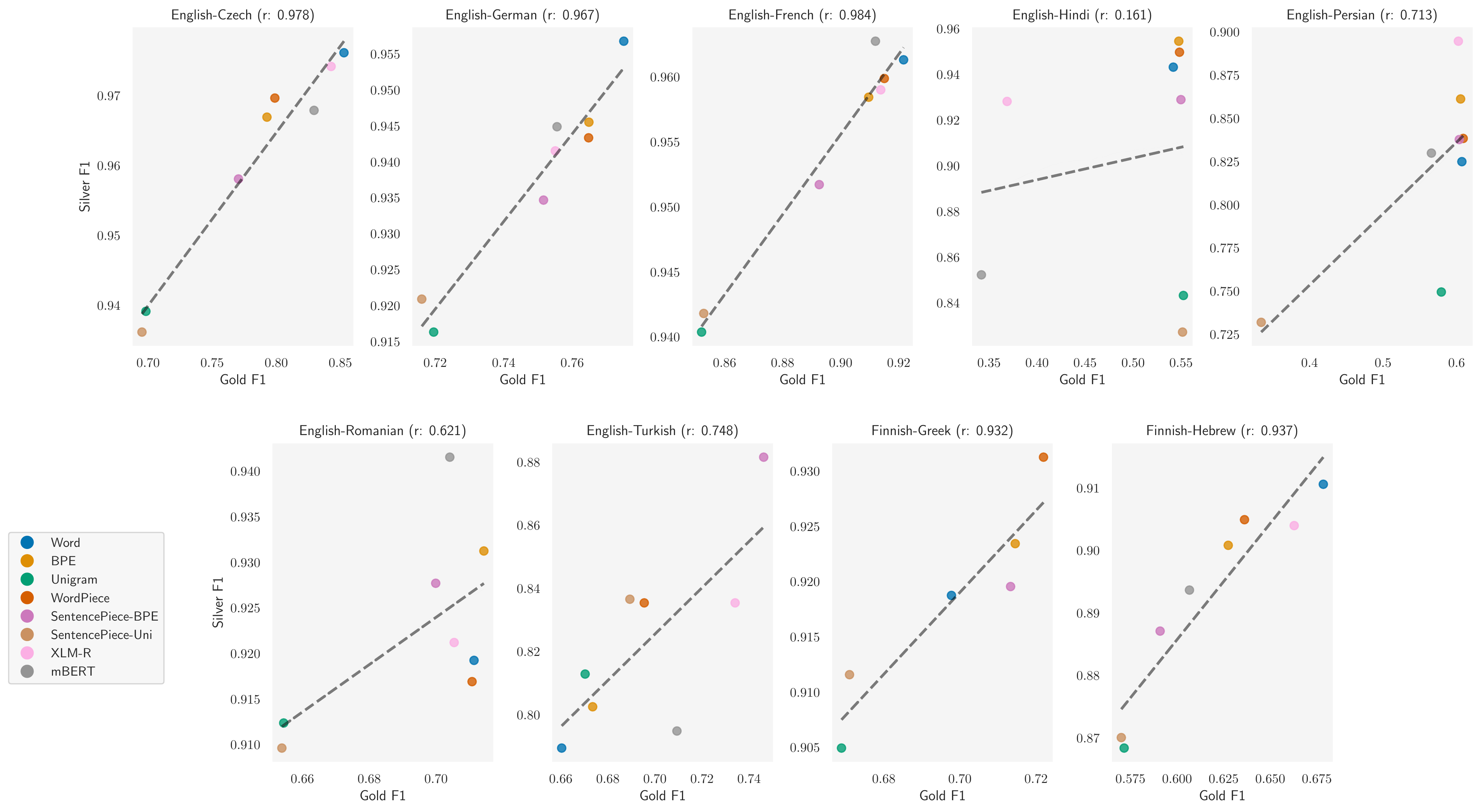}
		\caption{Comparison of tokenizer
			performance on Gold and
			Silver\textsubscript{Small}. Pearson's r correlation coefficient
			of $\fone$ scores is reported in the title
			of each language pair subfigure.
			The subfigures demonstrate that
			there is a strong correlation between Silver and Gold when different tokenizers are ranked.
			Thus, we can
			identify the best performing tokenizer based on Silver data if
			Gold data are not available. 
		}
		\label{fig:res_tokenizers}
	\end{figure*}

	\subsection{Evaluation}
	We compare word alignment for different setups with gold data and silver data to show the correlations between the latter. For word alignment, we report $\fone$ scores.\footnote{Alignment Error Rate (AER) and $\fone$ scores were calculated for all experiments. Since they led to similar conclusion, we only report $\fone$ for clarity.} For a given set of word alignment prediction edges $A$, a set of sure alignments $S$, and a set of possible gold (or silver) alignments $P$, we calculate the $\fone$ as:
	\begin{align*}
		&\fone = 2\frac{\precision\cdot \recall}{\precision+\recall},\\
		\text{where  } & \precision = \frac{|A \cap P|}{|A|}  \text{, } \recall= \frac{|A \cap S|}{|S|}\\
	\end{align*}
	
	Silver alignments are not complete, so we perform a
	partial evaluation: we only evaluate with a subset of
	predicted edges $A$ which include alignments $(s_i, t_j)$
	where $s_i$ is aligned to at least one word in the silver
	data. Thus, we expect higher $\fone$ scores for the silver
	datasets. Our goal in using silver datasets is not an
	accurate assessment of absolute $\fone$, but rather an
	assessment of
	relative performance, e.g., which tokenization method is better.
	
	For the verification of the silver data quality, we evaluate different sets of settings (e.g. tokenizers) and compare the correlation of scores for gold and silver data for each language pair. We report Pearson correlation coefficient (Pearson's $r$) \cite{freedman2007statistics}.

	\section{Results}
	\label{results}
	We conduct multiple experiments and evaluations to compare the performance of our silver sets and to analyze different tokenizers.

	\subsection{Silver Dataset}\label{sec:silver_dataset}
	
	We use gold and parallel data in
	Table \ref{tab:data} for our experiments. We apply \ourmethod for creating two silver sets for each language:
	Silver\textsubscript{Small} and Silver\textsubscript{Large}. 
	Silver \textsubscript{Small} uses the source sentences in the gold data (English or Finnish).
	As our method selects 
	safe alignments with more than three alternatives,
	it may produce a low number of total alignments and
	may not generate any alignments for some sentences
	as shown in Table \ref{tab:silver}. However, in the
	next sections, we will show that
	Silver\textsubscript{Small} demonstrates a strong
	correlation with the gold data even if the relative number of alignments is small.

	We introduce Silver\textsubscript{Large} to better illustrate our findings, especially when the gold sets contain less than 200 sentences. It is created by applying \ourmethod to 50,000 random English sentences from the C4 real news corpus \cite{raffel-etal-2020-exploring}.
	In section \ref{sec:pos}, we show that when the number of monolingual sentences is bigger, our method can find diverse alignment links in terms of frequency and PoS tags. We also
	generally 
	observe a strong correlation between Silver\textsubscript{Large} and Gold even though they are sampled from different domains highlighting the wide applicability of \ourmethod.
	In all the experiments that include
	Silver\textsubscript{Small} and
	Silver\textsubscript{Large}, we use the same
	parallel data for word alignment training, shown in Table \ref{tab:data}.

	\subsection{Subword Tokenization}\label{subword_tokenization}
	
	In our first experiment, we analyze the effects of subword tokenization. For a given language pair, we train a tokenizer with a shared vocabulary size 50,000 for source and target languages. 
	We compare Byte-Pair Encoding (BPE) \cite{gage1994new_bpe}, Unigram \cite{kudo2018subword_unigram}, WordPiece \cite{schuster2012japanese_wordpiece}, SentencePiece \cite{kudo2018sentencepiece} with Unigram, and SentencePiece with BPE tokenizers. We also include word-level tokenization and the tokenizer of mBERT \cite{devlin-etal-2019-bert} and XLM-R \cite{conneau-etal-2020-unsupervised}. 
	We use HuggingFace's tokenizer library for the implementation.\footnote{\url{https://github.com/huggingface/tokenizers}}
	Please note that the mBERT tokenizer contains an additional splitting mechanism (i.e., hyphen (-) and apostrophes (\textquotesingle)) on top of whitespace splitting and the XLM-R tokenizer includes tokens with only whitespace. We consider this
	when mapping tokens back to words because token ids do not match to index of a character sequence in a tokenized text, split from whitespace.
	Also, we do not include mBERT and XLM-R tokenizers
	for Finnish-Greek. The reason is that they support
	modern Greek but not Ancient Greek, and its use
	would cause a significant mismatch because of many
	unknown tokens (41\% vs 0\% [UNK] ratio) in mBERT
	and because of aggressive word splitting (3.12 vs. 1.64 token to word ratio) in XLM-R for Ancient Greek compared to modern Greek.
	
	Figure \ref{fig:res_tokenizers} shows tokenizer performance evaluated on Gold and Silver\textsubscript{Small}.
	Tokenizer rankings
	are strongly correlated for gold and silver,
	with a 0.86 average r score (except English-Hindi, probably due to the very low amount of Gold data (90 sentences). We also observe better correlation in language pairs with high-quality gold data. Gold in English-Czech, English-German, and English-French include possible alignments to take ambiguity of the word alignment task into account \cite{matusov2004symmetric}. Gold sets in Finnish-X tackle this problem by introducing large-scale gold datasets. We observe that Silver\textsubscript{Small} has 0.96 average r score with Gold in these five language pairs. This shows that \ourmethod automatically creates a benchmark that strongly correlates with high-quality gold datasets.

	Unigram and SentencePiece-Unigram have the worst
	performance among all tokenizers when compared on
	all language pairs.
	The  gold data of
	English-Hindi are an exception; however,
	the silver data
	of English-Hindi align well with these pattern
	across all languages.
	
	We find that word tokenization is already strong for many languages such as English-Czech, English-German, and English-French. However, we see that subword level tokenization outperforms word tokenization for English-Turkish and Finnish-Greek. 
	In all experiments, XLM-R and mBERT tokenizers achieve comparable results to custom subword tokenization and word tokenization. 
	
	Comparisons of Gold with Silver\textsubscript{Large}
	are in the Appendix,
	Figure \ref{fig:res_tokenizers_c4}. We find similar
	performance as Silver\textsubscript{Small} even
	though the larger sets come from C4 real news
	corpus, a domain different from the domain of Gold
	(i.e., there is a domain adaptation aspect to this evaluation).
	
	In the rest of the results section, we use
	Silver\textsubscript{Large} for English-X and
	Silver\textsubscript{Small} for Finnish-X pairs to
	perform a more granular analysis.\footnote{Note that
		there is no Silver\textsubscript{Large} for
		Finnish-X since the small sets already contain more
		than 7K sentences. That is, for Finnish-X the
		``small'' sets are in reality large.}

	\subsection{Tokenizer Vocabulary Size}\label{sec:tokenizer_vocabulary_size}
	\label{sec:silver_data_size}
	
	\begin{figure*}[t]
		\centering
		\includegraphics[width=\textwidth]{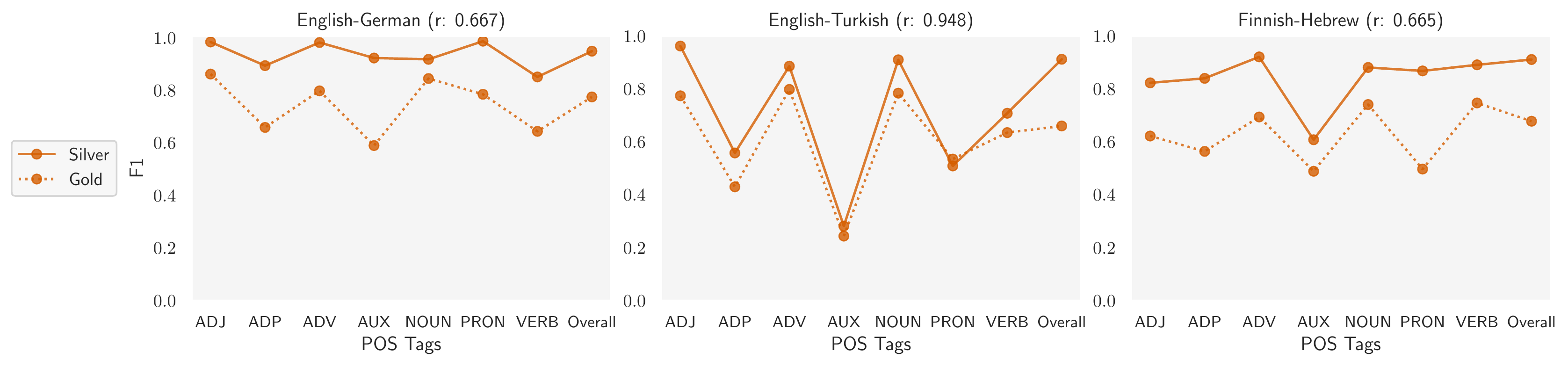}
		\caption{Analysis of word alignment
			performance with word-level tokenization
			for different part-of-speech tags.
			The title for each subfigure gives correlation
			(Pearson's r) of Silver
			and Gold $\fone$ across the different
			part-of-speech tags. 
			This shows that Silver is able to capture the relative performance of a word alignment method in PoS tags, similarly to Gold.
		}
		\label{fig:res_pos_size_r}
	\end{figure*}
	\begin{figure}[!h]
		\centering
		\includegraphics[width=0.4\textwidth]{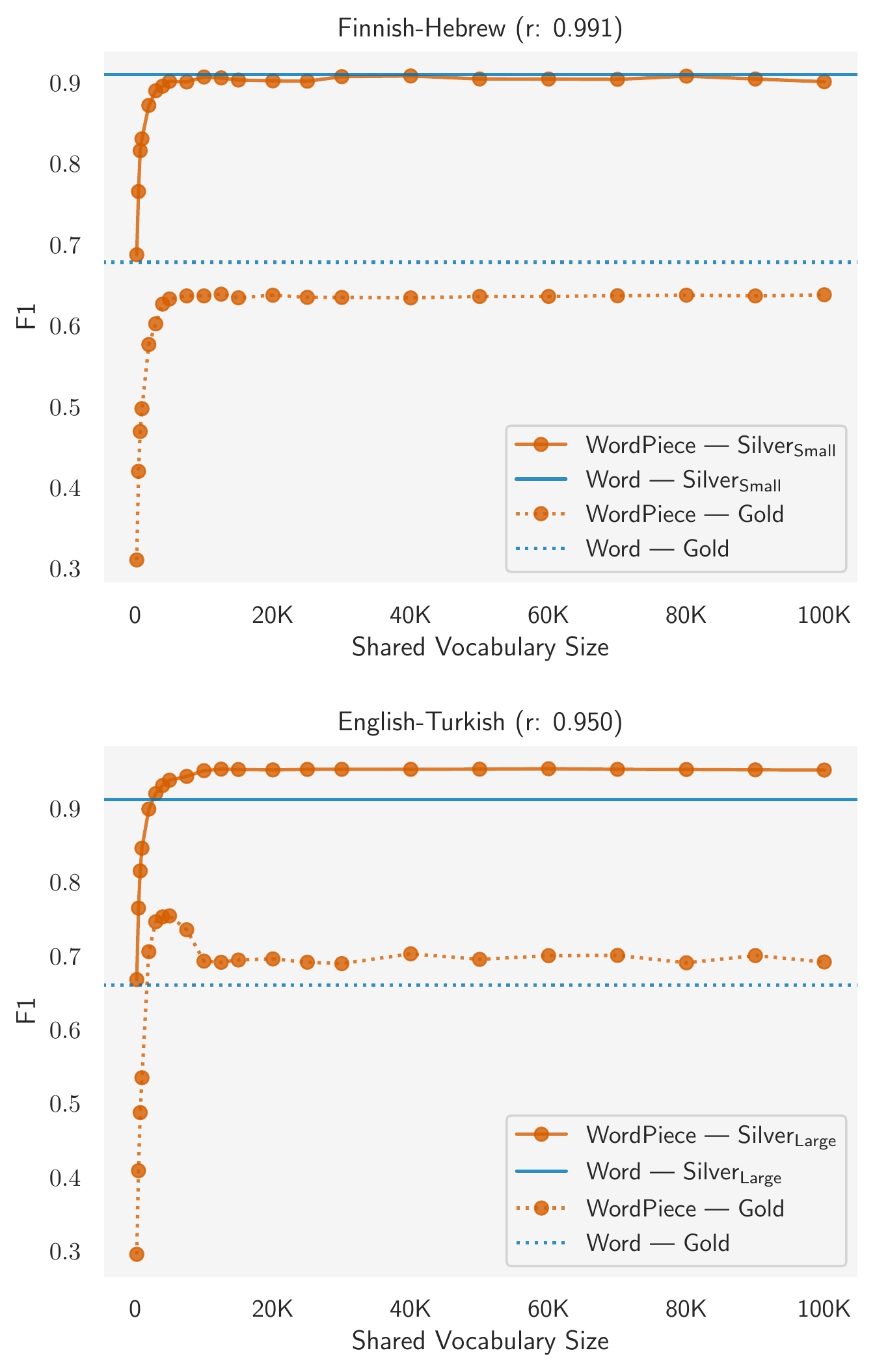}
		\caption{Comparison between word and WordPiece tokenizers with varying vocabulary sizes. 
			The figure shows that there is a
			strong correlation (Pearson's r, reported in the
			title of each subfigure) between $\fone$ scores of Silver and Gold when word and WordPiece tokenizers with varying number of vocabulary sizes are compared. Therefore, Silver can find the best vocabulary size for a given language pair with a high correlation with Gold.
		}
		\label{fig:res_tokenizer_vocab}
	\end{figure}

	As the tokenizer's vocabulary size is an important aspect in word alignment \cite{ho2021optimizing}, we experiment with different settings with shared vocabulary size. We use a shared WordPiece tokenizer and illustrate changes in $\fone$ score at varying vocabulary sizes. Results are depicted in Figure \ref{fig:res_tokenizer_vocab}.
	For both silver and gold, there's a strong correlation between word and WordPiece tokenization and vocabulary size change in WordPiece with an average 0.96 r score in all language pairs. We observe that WordPiece performs better than word tokenization for English-Turkish and vice versa for Finnish-Hebrew. 
	$\fone$ scores converge after around 10K vocabulary size for both language and dataset. There is a slight bump in Gold for English-Turkish around 4K vocabulary size. Since we do not observe a similar pattern in any language pairs in the gold data (see full plots in 
	Figure \ref{fig:res_tokenizers_c4}), we conjecture that the bump is due to the small amount of Gold data in Turkish.

	\subsection{Part-of-Speech Tagging Performance}\label{sec:pos}

	Previous works \cite{sabet2020simalign, ho2019neural} show that word aligners' performance can significantly vary with respect to different parts of speech. 
	Therefore, we investigate this aspect of our algorithm. 
	We use the Stanza \cite{qi-etal-2020-stanza} toolkit to tag the source sentences in English and Finnish. We compare Gold and Silver performance for different PoS tags (Silver\textsubscript{Large} for English-X and Silver\textsubscript{Small} for Finnish-X). 
	
	Figure \ref{fig:res_pos_size_r} shows similar
	patterns and high correlation between gold and
	silver.  For example, auxiliaries and adpositions
	perform significantly worse than other PoS tags in
	English-Turkish. This is because Turkish is an
	agglutinative language in which adpositions are usually case
	marked in noun forms, and auxiliaries are
	represented in suffixes in verb complexes. Also,
	word ordering on the English side is not
	monotonically aligned with the morpheme order of the
	Turkish counterpart of English adpositions and
	auxiliaries \cite{el2009exploiting} which makes
	accurately aligning words with those tags more
	difficult.
	
	\begin{figure}[t!]
		\centering
		\includegraphics[width=0.45\textwidth]{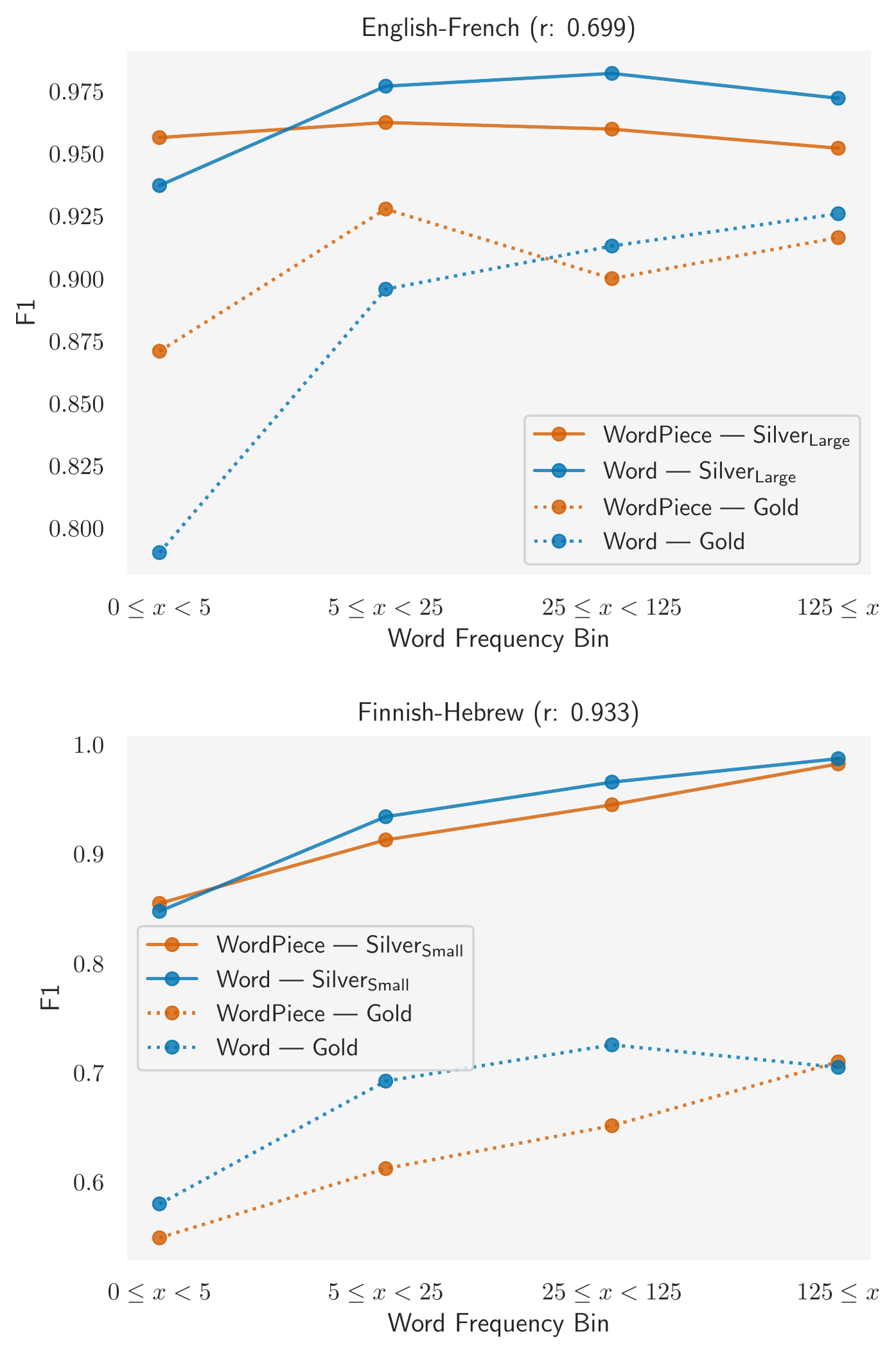}
		\caption{Comparison of the word and WordPiece tokenization with respect to $\fone$ scores of different word frequency bins on Silver and Gold. This figure demonstrates that similar performance's patterns on different frequencies and tokenizers can be observed between Silver and Gold.}
		\label{fig:res_word_frequency}
	\end{figure}
	This experiment also illustrates that a silver dataset created with \ourmethod contains a diverse set of word alignments to infer additional information about linguistic properties. We present PoS distribution with respect to gold and silver for all language pairs in the Appendix, Figure \ref{fig:res_pos_size_r_full}.

	\subsection{Word Frequency}\label{sec:word_frequency}
	
	We compare word and WordPiece tokenizations for
	different word frequency bins in
	Figure \ref{fig:res_word_frequency}.  The frequency
	of a word is defined as the minimum frequency of
	source and target words for a predicted alignment.
	Similar to \citet{sabet2020simalign}, we observe
	that subword tokenizations, like WordPiece, perform
	better than word tokenization on low-frequency words
	for English-French while we do not observe such
	major performance difference for Finnish-Hebrew in
	Figure \ref{fig:res_word_frequency}. Even though
	word and WordPiece tokenizers perform comparably for
	these languages, we observe that the impact on low-
	and high-frequency words might be quite
	different. Therefore, tokenizers can be selected
	according to sub-objectives by using Silver data,
	obviating the need for creating an expensive gold
	data benchmark. We present word and WordPiece performances on
	word frequency for all language pairs in the
	Appendix, Figure \ref{fig:res_word_frequency_full}.

	\section{Conclusion}
	\label{conclusion}
	Since creating a human-annotated word alignment dataset is a challenging task, we propose the \ourmethod method to create a silver benchmark using a Masked Language Model (MLM) and a machine translation model. 
	\ourmethod makes use of MLMs to create minimal pairs with alternatives that fit well into context and find partial alignments based on the changes in the translation of the alternatives via machine translation. 
	
	We show that our method can create a high-quality
	silver benchmark for 9 language pairs including
	pairs of two non-English languages. We show that the silver benchmark on two different domains (Silver\textsubscript{Small} and Silver\textsubscript{Large}) can help to compare different configurations and investigate errors with a high correlation to the gold data. We perform experiments on sub-word level tokenization, tokenizer vocabulary size, and performance change with respect to PoS tags and word frequency. 
	
	For future work, \ourmethod can be extended to create a specific subset of a general domain dataset to analyze the effects of potential issues in word alignment such as rare words. We believe that \ourmethod can ease up the process of finding issues in existing word alignment models for various language pairs, and it can help to improve both word alignment tools and tasks that use word alignment implicitly or explicitly such as machine translation.
	
	Finally, we believe that our silver data creation algorithm can be helpful for both low- and high-resource language pairs to investigate word alignment without 	a
	time-consuming human annotation process. If combined
	with recent machine translation models
	(e.g. NLLB \cite{nllb}), \ourmethod can, in
	principle, support more than 200 languages.
	Therefore, we make our silver data and code available as a
	resource for future work that takes advantage of our silver evaluation datasets.\footnote{\url{https://github.com/akoksal/SilverAlign}}

	\section*{Limitations}
	\ourmethod is limited to language pairs with existing machine translation systems and MLMs for the source language. 
	Even though there are recent works and commercial tools that support hundreds of languages for machine translation and MLM, the quality of these systems should be taken into account. 
	For low-quality MT and MLM systems, \ourmethod might require larger monolingual corpora in the source language to create a silver dataset with a good amount of total alignment.
	
	For evaluation purposes, we only evaluate SilverAlign on language pairs for which gold alignments are available. However, our method is applicable to any language pairs for which MT systems and MLMs are available for. Therefore, this includes languages that are more ``low-resource'' with respect to the one addressed in this paper.

	\section*{Ethics Statement}
	
	Our work is based on publicly available datasets. We
	would like to clarify that we treat the data simply
	to prove our points, and the content does not
	necessary reflect the authors' opinions or the
	opinions of their funding institutions.
	
	\section*{Acknowledgements}
	This work was funded by the European Research Council (grant \#740516) and the German Federal Ministry of Education and Research (BMBF, grant \#01IS18036A).

	\bibliographystyle{acl_natbib}
	\bibliography{arxiv_silveralign}
	
	\appendix
	\begin{figure*}[ht!]
		\centering
		\includegraphics[width=\textwidth]{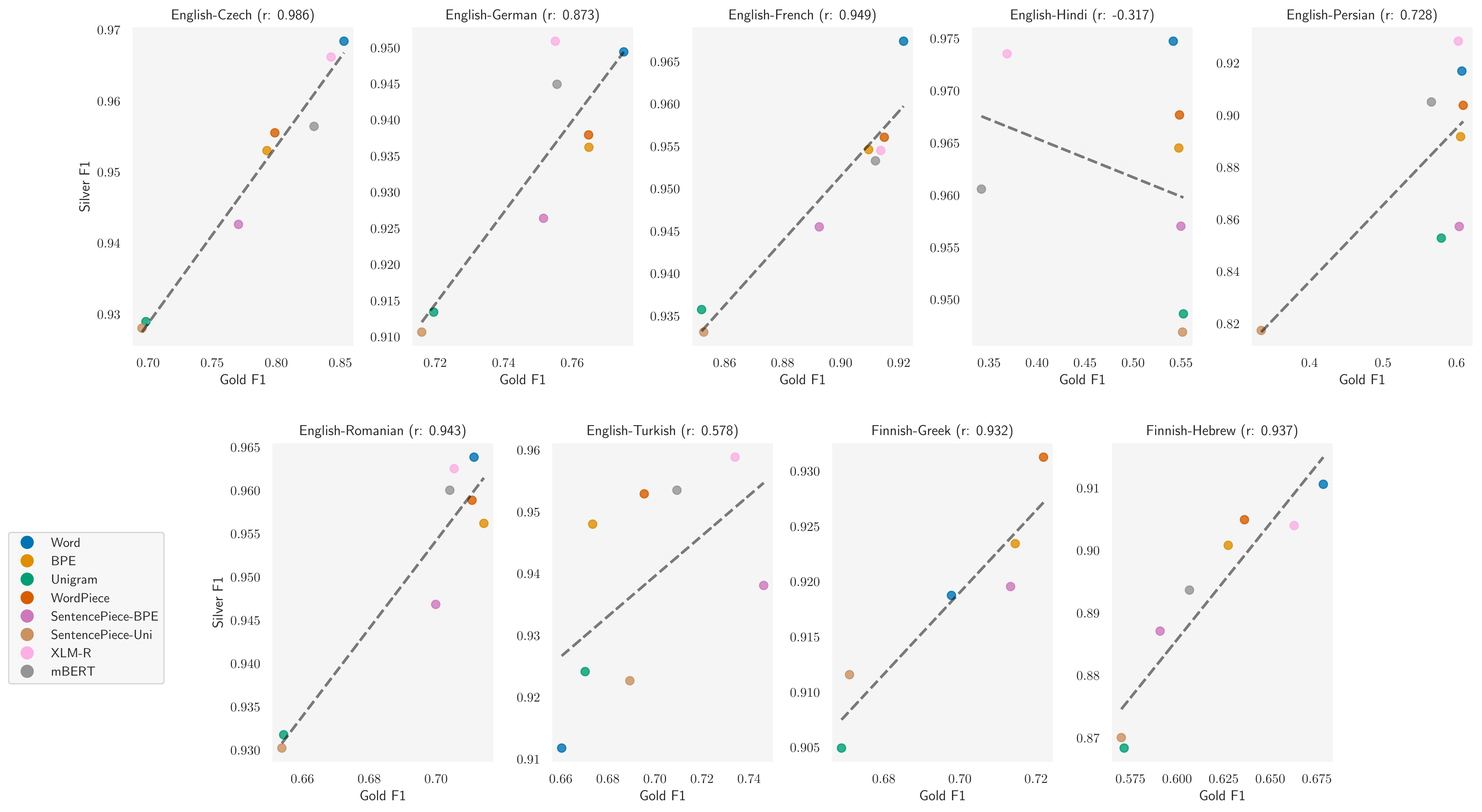}
		\caption{Comparison of tokenizer
			performance on Gold and
			Silver\textsubscript{Large} (Silver\textsubscript{Small} for Finnish-X). 
			The Pearson's r
			of $\fone$ scores are reported in the title
			of each language pair subfigure.
			The figures demonstrate that there's a strong correlation between Silver and Gold when different tokenizers are ranked. 
			Thus, we can
			identify the best performing methods based on Silver data if
			Gold data are not available. }
		\label{fig:res_tokenizers_c4}
	\end{figure*}

	\begin{figure*}[ht!]
		\centering
		\includegraphics[width=\textwidth]{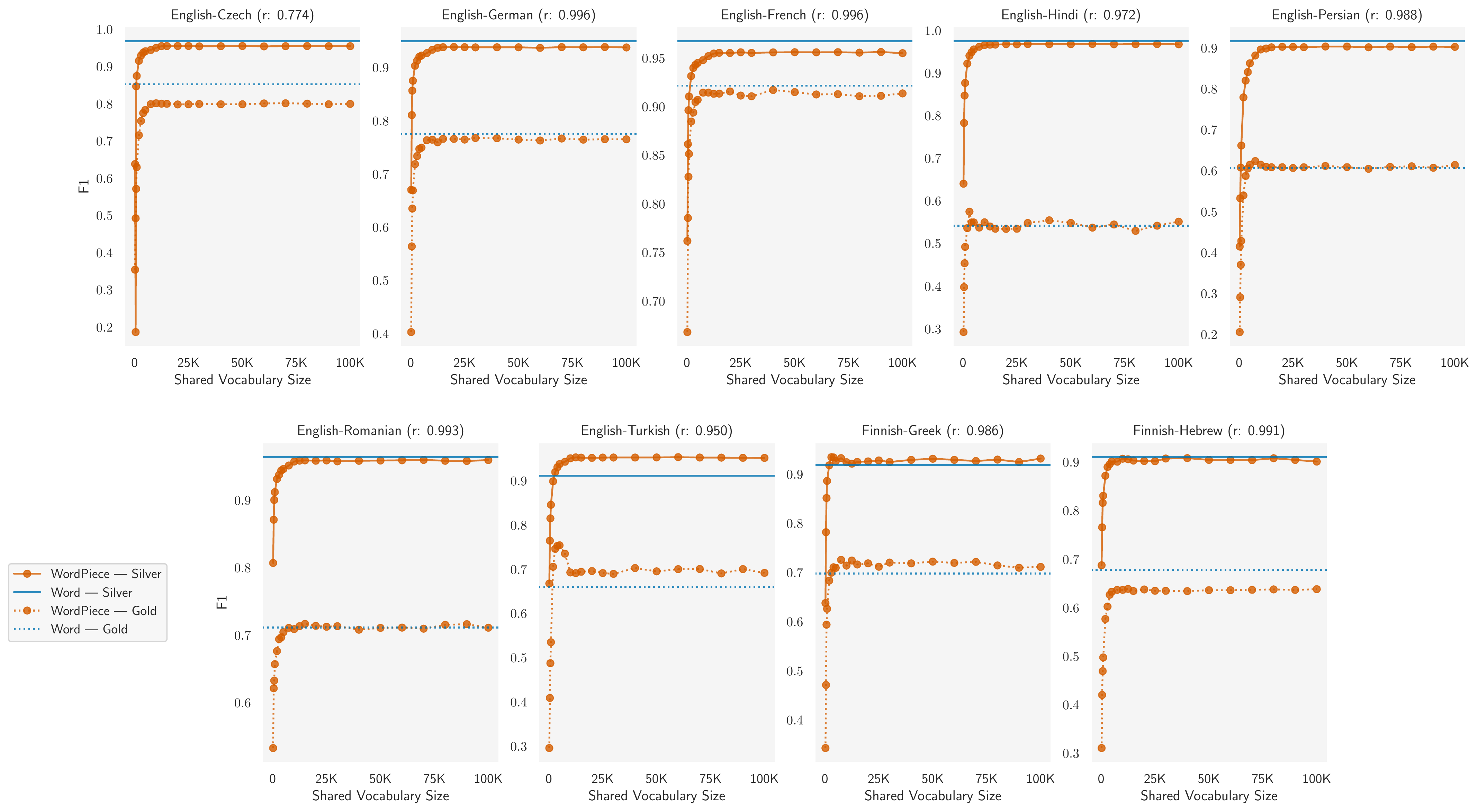}
		\caption{Comparison between word and WordPiece tokenization with varying vocabulary sizes. 
			The figure shows that there's a strong correlation (Pearson's r) between $\fone$ scores of Silver and Gold when word and WordPiece tokenizer with varying number of vocabulary size are compared, as reported in the title. Therefore, Silver can find the best vocabulary size for a given language pair with a high correlation with Gold.}
		\label{fig:res_tokenizer_vocab_full}
	\end{figure*}

	\begin{figure*}[ht!]
		\centering
		\includegraphics[width=\textwidth]{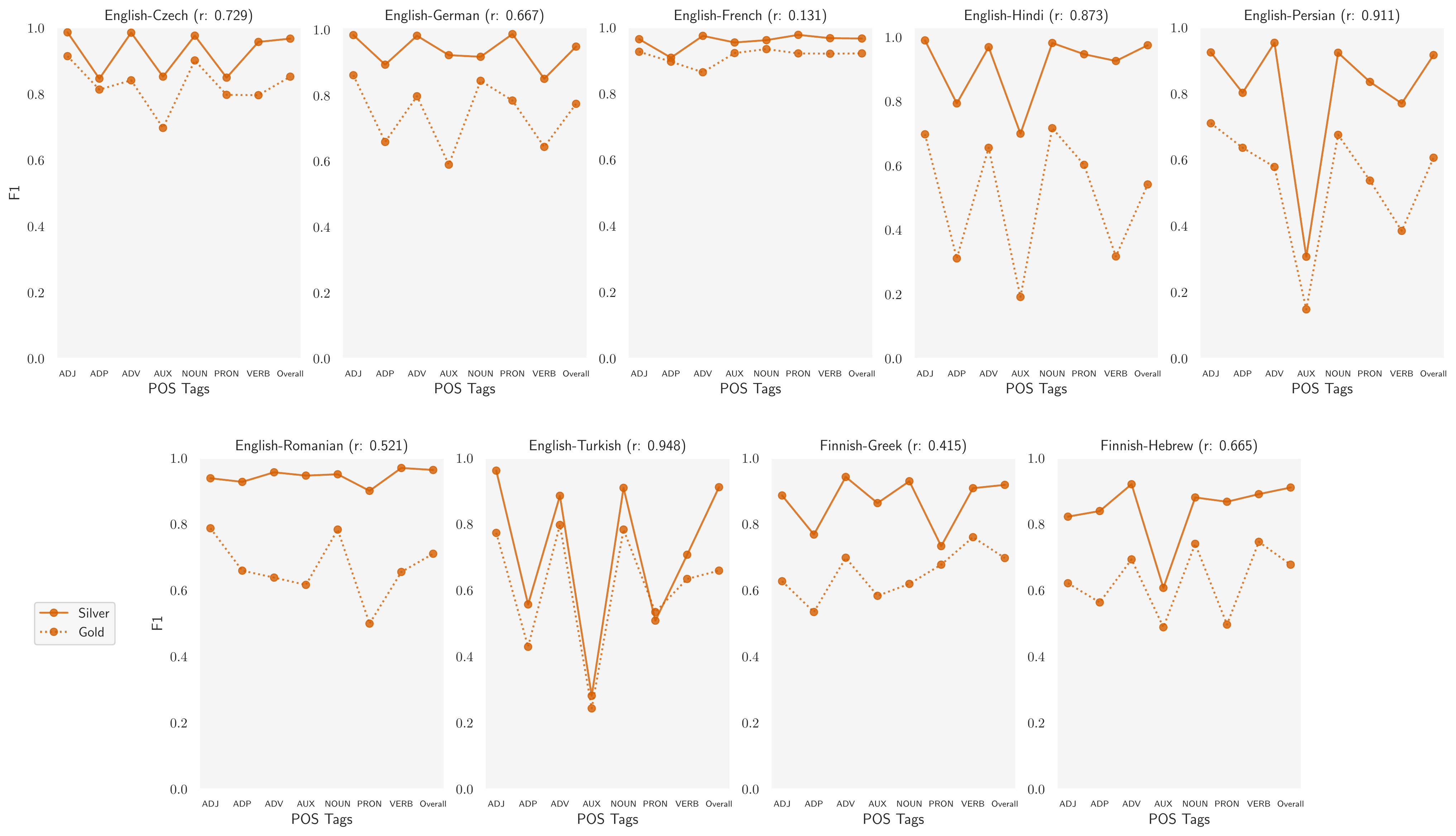}
		\caption{Analyzing word alignment
			performance with word-level tokenization
			for different part-of-speech tags.
			The title for each subfigure gives correlation
			(Pearson's r) of Silver
			and Gold $\fone$ across the different
			part-of-speech tags. 
			This shows that Silver is able to capture the relative performance of a word alignment method in PoS tags, similarly to Gold. 
		}
		\label{fig:res_pos_size_r_full}
	\end{figure*}
	
	\begin{figure*}[ht!]
		\centering
		\includegraphics[width=\textwidth]{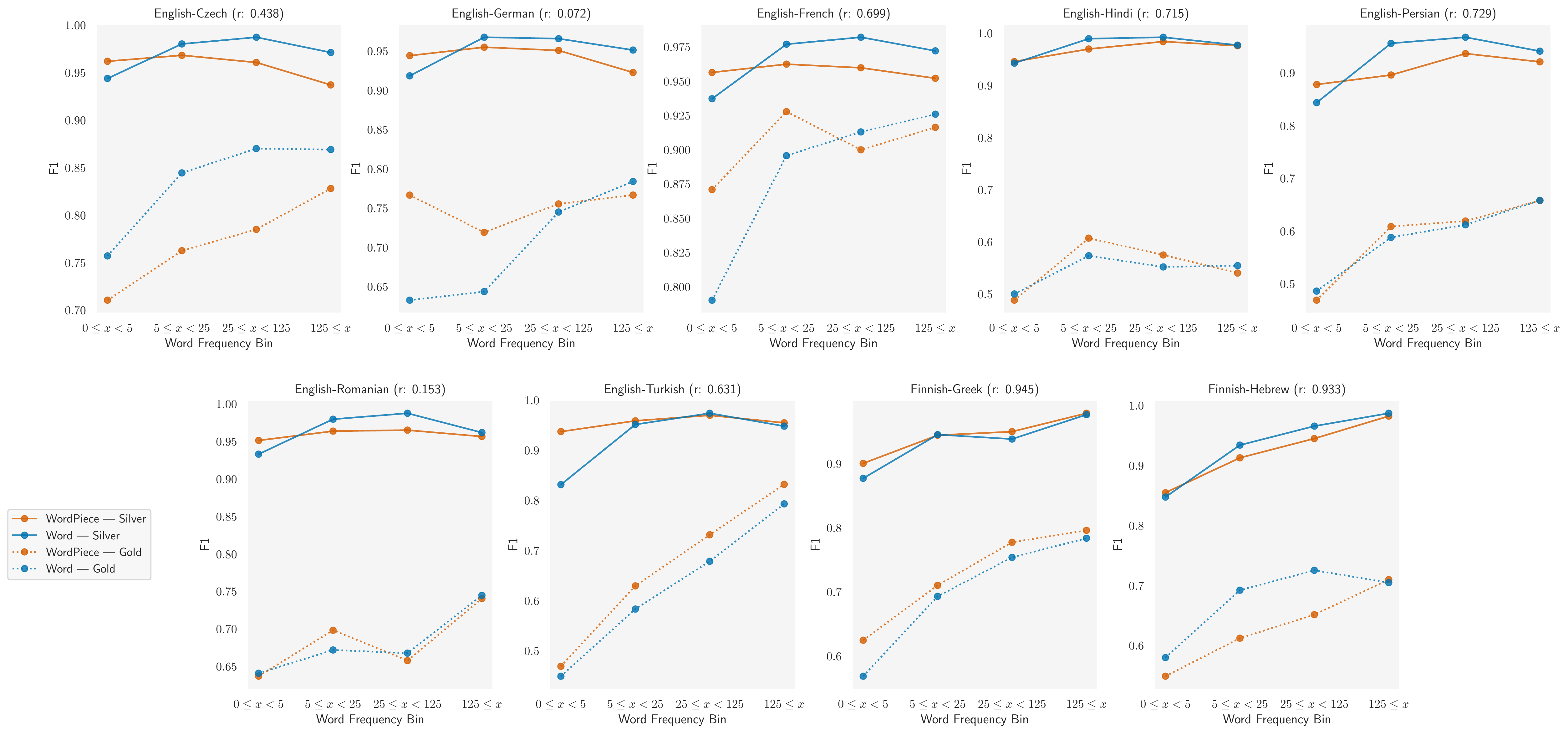}
		\caption{Comparison of the word and WordPiece tokenization with respect to $\fone$ scores of different word frequency bins on Silver and Gold. This figure demonstrates that similar performance's patterns on different frequencies and tokenizers can be observed between Silver and Gold.}
		\label{fig:res_word_frequency_full}
	\end{figure*}

\end{document}